\documentclass[10pt,twocolumn,letterpaper]{article} 

\usepackage{bm}
\usepackage{lineno}
\usepackage[pagenumbers]{cvpr} 
\usepackage[utf8]{inputenc}

\newcommand{\cyan}[1]{\textcolor{cyan}{#1}}

\definecolor{cvprblue}{rgb}{0.21,0.49,0.74}
\usepackage[pagebackref,breaklinks,colorlinks,allcolors=cvprblue]{hyperref}

\def\paperID{9814} 
\def\confName{CVPR}
\def\confYear{2026}

\title{VibraVerse: A Large-Scale Geometry-Acoustics Alignment Dataset for Physically-Consistent Multimodal Learning}

\author{Bo Pang\\
Peking University\\
Beijing, China\\
{\tt\small bo98@stu.pku.edu.cn}
\and
Chenxi Xu\\
Peking University\\
Beijing, China\\
{\tt\small xuchenxi@pku.edu.cn}
\and
Jierui Ren\\
Peking University\\
Beijing, China\\
{\tt\small jerry@stu.pku.edu.cn}
\and
Guoping Wang\\
Peking University\\
Beijing, China\\
{\tt\small wgp@pku.edu.cn}
\and
Sheng Li\\
Peking University\\
Beijing, China\\
{\tt\small lisheng@pku.edu.cn}
}

\begin{document}
\twocolumn[{%
 \renewcommand\twocolumn[1][]{#1}%
 \maketitle
 \includegraphics[width=.99\linewidth,trim=0 2cm 0 0,clip]{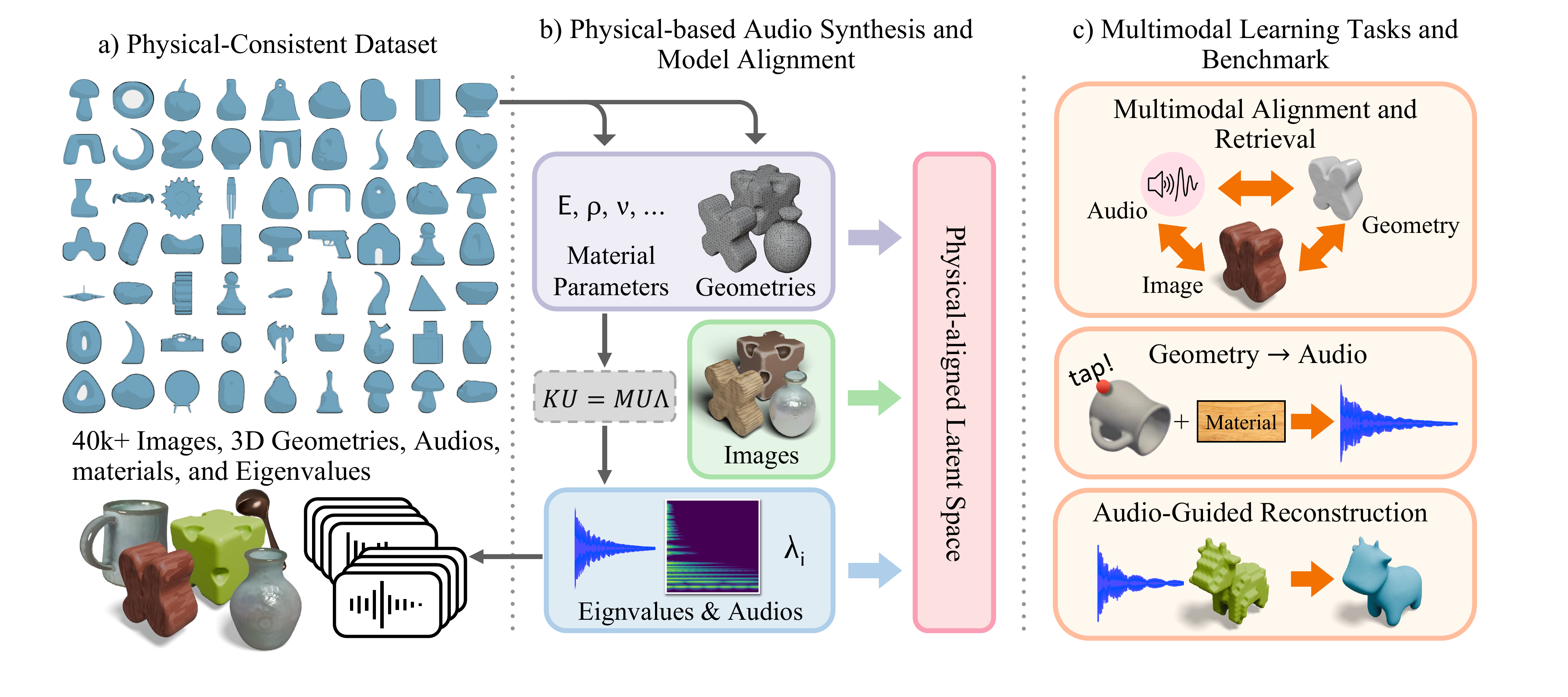}
 \captionof{figure}{Overview of our framework for physically-consistent geometry–acoustics learning.
 (a) We build a large-scale physically-consistent dataset comprising over 40K objects, each annotated with images, 3D geometries, materials, eigenvalues, and physically synthesized audios. 
 All data are generated under unified physical parameters to ensure geometry–material–acoustics consistency. 
 (b) Using finite-element modal analysis, we derive eigenfrequencies and modal sounds, aligning each object’s geometry and material with its intrinsic acoustic response in a shared physics-grounded latent space. 
 (c) This physically-consistent dataset serves as the foundation for multimodal learning and reasoning, enabling 
 \textbf{cross-modal alignment and retrieval}, 
 \textbf{geometry-to-audio synthesis}, and 
 \textbf{audio-guided 3D reconstruction}. 
 The dataset and inference tasks establish a benchmark for physically-grounded multimodal understanding and sound-driven 3D reasoning, as a bridge enabling physically interpretable multimodal understanding of the physical world.}
 \vspace{0.2in}
 \label{fig:teaser}
 }]

 \begin{abstract}
Understanding the physical world requires perceptual models grounded in physical laws rather than mere statistical correlations. However, existing multimodal learning frameworks, focused on vision and language, lack physical consistency and overlook the intrinsic causal relationships among an object’s geometry, material, vibration modes, and the sounds it produces.
We introduce VibraVerse, a large-scale geometry–acoustics alignment dataset that explicitly bridges the causal chain from 3D geometry → physical attributes → modal parameters → acoustic signals. Each 3D model has explicit physical properties (density, Young’s modulus, Poisson’s ratio) and volumetric geometry, from which modal eigenfrequencies and eigenvectors are computed for impact sound synthesis under controlled excitations.
To establish this coherence, we introduce CLASP, a contrastive learning framework for cross-modal alignment that preserves the causal correspondence between an object’s physical structure and its acoustic response.
This framework enforces physically consistent alignment across modalities, ensuring that every sample is coherent, traceable to the governing equations, and embedded within a unified representation space spanning shape, image, and sound.
Built upon VibraVerse, we define a suite of benchmark tasks for geometry-to-sound prediction, sound-guided shape reconstruction, and cross-modal representation learning.
Extensive validations on these tasks demonstrate that models trained on VibraVerse exhibit superior accuracy, interpretability, and generalization across modalities.
These results establish VibraVerse as a benchmark for physically consistent and causally interpretable multimodal learning, providing a foundation for sound-guided embodied perception and a deeper understanding of the physical world.
The dataset will be open-sourced.

\end{abstract}    
 \section{Introduction}
\label{sec:intro}


Sound and geometry are inherently linked through the laws of physics: when an object vibrates or is struck, its shape and material properties jointly determine how it resonates and emits sound. In essence, sound is the temporal and spectral projection of an object’s geometry and physical constitution. Humans naturally leverage this relation, and we can often infer an object’s material or thickness simply from the way it sounds. However, such auditory-based reasoning remains largely unexplored.

Early works in computer graphics and computational acoustics \cite{van2001foleyautomatic,chadwick2009harmonicshells} showed that an object’s eigenfrequencies and eigenmodes can be derived from its geometry and material properties via finite-element or modal analysis.
However, these physically based methods function only as forward models for sound synthesis, unable to address the inverse problem of inferring geometry from sound.

In computer vision and multimodal learning, several recent works have attempted to connect auditory and visual cues. For instance, \textit{SoundSpaces}~\cite{chen2020soundspaces}, and \textit{MultiFoley}~\cite{chen2025video}, explored linking sound with visual scenes or coarse 3D reconstruction from videos. Yet, these datasets are based on real-world recordings, which suffer from uncontrolled excitation, environmental noise, and unknown material properties, thereby lacking physical consistency between geometry and sound.  
Recently, DiffSound~\cite{Jin2024DiffSound} introduced a differentiable modal sound rendering framework that enables inverse inference of geometry and material from sound under a fully physics-based pipeline.

From a data perspective, existing large-scale datasets focus primarily on semantics rather than physical causality. Audio–visual datasets such as \textit{AudioSet}~\cite{gemmeke2017audioset}, \textit{VGGSound}~\cite{chen2020vggsound}, \textit{Fair-Play}~\cite{gao2018learning}, and \textit{SoundSpaces}~\cite{chen2020soundspaces} capture environmental or human-generated sounds but omit object-level geometry and material information. Conversely, 3D geometry datasets such as \textit{ShapeNet}~\cite{chang2015shapenet}, \textit{ModelNet}~\cite{wu2015modelnet}, and \textit{Objaverse}~\cite{deitke2023objaverse} contain rich shape diversity but lack any acoustic or modal annotations related to physical attributes. 


While the ObjectFolder series \cite{gao2021ObjectFolder,gao2022objectfolder,Gao_2023_CVPR} advanced audio related multisensory learning, it remains limited in capturing physically grounded geometry–acoustics relationships.
\textbf{First}, the coupling among geometry, material, and sound is implicit rather than causal: auditory signals are produced from event-based surface interactions and generally represent empirical correlations between modalities but lacking explicit modeling of the physical mechanisms that link shape and material to acoustic behavior.
\textbf{Second}, current datasets omit physically parameterized representations, including volumetric geometry, modal spectra, eigenfrequencies, and material-dependent acoustic properties necessary for describing intrinsic physical characteristics and enabling cross-domain generalization beyond purely data-driven associations.
\textbf{Finally}, there is no systematic benchmark for assessing geometry–acoustics consistency or for evaluating physically grounded reasoning tasks, such as sound-guided shape reconstruction, geometry-to-sound synthesis, and material inference, that require causal, physically interpretable understanding.

The absence of explicit physical grounding constrains representation learning, generalization, and interpretability.
Without physical parameters, models capture only statistical correlations rather than the causal mechanisms governing real-world object behavior, leading to poor generalization across variations in shape, material, and volumetric structure.
Incorporating physical features derived from modal analysis enables physically interpretable multimodal learning, improved cross-domain generalization, and quantitative evaluation of geometry–acoustics consistency.

\begin{figure}[htbp]
  \centering 
  \includegraphics[width=0.85\linewidth]{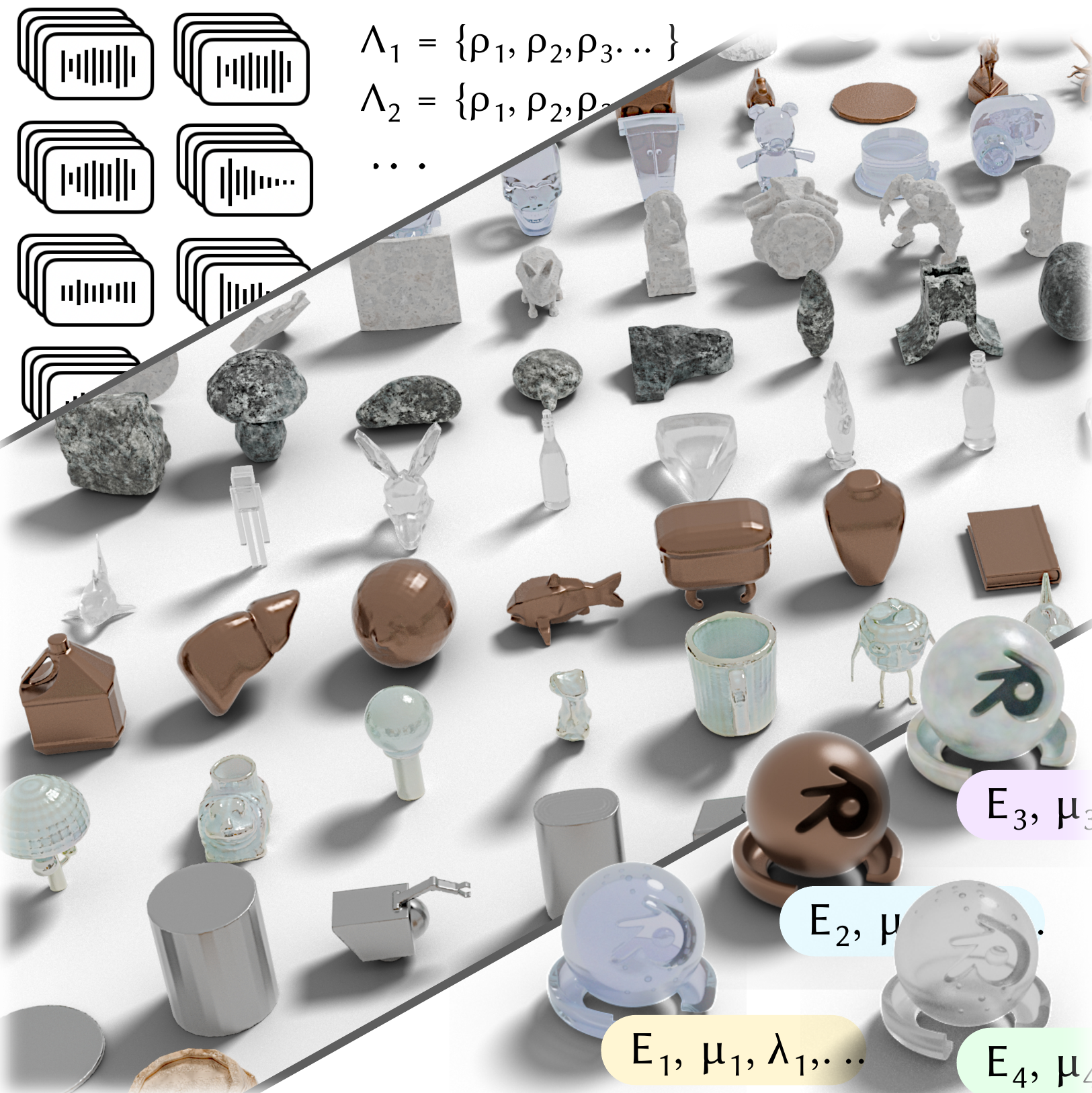}
  \caption{
  The VibraVerse dataset comprises a diverse collection of objects spanning a wide range of physical materials (bottom). Each object is defined by its physical parameters, which are utilized to synthesize corresponding eigenfrequencies, eigenmodes, and modal sounds (top). This process establishes a physically grounded correspondence linking object geometry, material properties, and acoustic signatures.}
  \label{fig:dataset_demo}
\end{figure}

\textbf{Motivation and Objectives}: To bridge these gaps, we aim to construct a \textbf{physically consistent geometry–acoustics alignment dataset} that explicitly encodes the causal chain:
\text{Geometry (3D shape and volumetric data)} $\rightarrow$ Physical properties (E, $\rho$, $\nu$) $\rightarrow$ Modal parameters (eigenvalue, eigenvector) $\rightarrow$ Sound signals.
Each 3D model has explicit material parameters (density, Young’s modulus, Poisson’s ratio) and is subjected to modal analysis to compute its vibration modes. The resulting eigenfrequencies and mode shapes are then used to synthesize the corresponding impact sound under controlled excitation. This process guarantees that every sample in the dataset is physically coherent, where geometry, material, and sound are tightly coupled through physical laws and fully traceable to simulation parameters.  

Building upon this foundation, we introduce a large-scale dataset for physically consistent multimodal learning, enabling AI systems to jointly reason across 3D geometry, 2D image, material, and impact sound.
We further design a suite of novel cross-modal and inverse reasoning tasks, including geometry-to-sound synthesis, sound-guided shape reconstruction, material identification, and tri-modal retrieval, many of which have not been feasible with previous datasets.
Together, these components provide a platform for evaluating physics-aware multimodal learning and establish a benchmark for physically grounded perception and reasoning beyond purely semantic or visual alignment (see Fig.\ref{fig:teaser}).


Our key contributions include:
\begin{enumerate}
    \item \textbf{A large physically-consistent geometry–acoustics dataset.}  Each object is associated with complete 3D geometry (both surface and volumetric data), physical attributes, modal spectra, and synthesized sound, forming a traceable physical chain. All samples have been verified for physical plausibility and simulation consistency.  
    \item \textbf{Multimodal tasks and benchmark.} We establish a suite of tasks, including \textit{geometry-to-sound prediction}, \textit{shape reconstruction}, \textit{cross-modal retrieval}, and \textit{material classification}, along with evaluation protocols and a physically aligned contrastive learning framework that unifies geometry–sound representations. Together, these form a benchmark for consistent evaluation of physically grounded multimodal reasoning.
    
\end{enumerate}

 \section{Related Works}
\label{sec:related_works}

\subsection{3D Datasets on Object-level Geometry}
With the rapid development of 3D vision tasks, numerous datasets and benchmarks have been proposed.
Early datasets such as ModelNet~\cite{wu2015modelnet} and ShapeNet~\cite{chang2015shapenet} provided large-scale collections of 3D models for object classification and segmentation tasks. 
Subsequently, datasets like Thingi10K~\cite{zhou2016thingi10k} and ABC~\cite{Koch2019} introduced more diverse and complex 3D shapes, enabling advancements in shape analysis and reconstruction. 
Recent web-scale datasets such as Objaverse~\cite{deitke2023objaverse}, Objaverse-XL~\cite{deitke2023objaverseXL}, Animal3D~\cite{xu2023animal3d}, and OmniObject3D~\cite{wu2023omniobject3d} further expand category coverage, spanning man-made, organic, and articulated objects.


\subsection{Acoustic-Related Datasets}
Several sound-related datasets have been developed to support research in audio recognition and generation.
AudioSet~\cite{gemmeke2017audio} serves as a foundational dataset for audio classification tasks, recording 632 categories of human-labeled sound events.
Zhang et al.~\cite{zhang2017generative} introduced a synthetic dataset of object shape, material and corresponding sound, enabling the study of sound generation from visual inputs.
Gao et al.~\cite{gao2022ObjectFolderV2, gao2022objectfolder} proposed ObjectFolder, an object-centric dataset of around 1,000 everyday objects with high-quality 3D models and sound simulations, together with touching events. 
Clarke et al.~\cite{clarke2023realimpact} presented RealImpact, which records 150k knock sound from 50 real-world objects in highly controlled acoustic environments.

\subsection{Multimodal Learning involving Physical Attributes}

Recent multimodal approaches increasingly incorporate \emph{physical attributes} such as material and contact dynamics, where impact acoustics provide cues complementary to visual appearance~\cite{clarke2023realimpact}. 
At the object level, Object-centric datasets such as ObjectFolder~\cite{gao2021ObjectFolder, gao2022ObjectFolderV2} capture multimodal data including geometry, rendering, and contact-induced sounds.
At the scene level, SoundSpaces~\cite{chen2020soundspaces, chen2022soundspaces} simulates room impulse responses aligned with 3D indoor environments.
The physical realism has been further improved by recent efforts in real-scene acoustic field measurements~\cite{chen2024real}
Other works like Neural Acoustic Fields and audio-visual neural radiance field models~\cite{luo2022learning, liang2023av} jointly encode geometry and sound propagation. 
However, these scene-level and acoustic-level approaches generally lack object-specific modal alignment. 

\section{VibraVerse Dataset}


In this section, we detail the construction of the VibraVerse dataset, a dataset including more than 40,000 3D objects and their corresponding physical properties, modal parameters, and synthesized sounds. 
A visual overview of the dataset is shown in ~\cref{fig:dataset_demo}.
We first provide an overview of modal analysis and sound synthesis.
As compact descriptors of an object's global physical behavior and material attributes, modal representations guide the data generation process and underpin our pursuit of physically consistent multimodal learning.
Then, we formally define the formulation of our dataset and its components.
Then, we detail the steps to create the 3D geometry, assign material properties, and perform acoustic simulation to generate physically consistent sound signals.

\subsection{Formulation of Modal Analysis and Sound Synthesis} \label{sec:method_background}

\paragraph{Modal Analysis.}

The synthesis of physically plausible impact sounds from 3D objects is fundamentally based on analyzing their inherent vibrational properties.
Given a 3D object represented by a volumetric mesh, its physical behavior is governed by its geometry and material properties, specifically density ($\rho$), Young's modulus ($E$), and Poisson's ratio ($\nu$). Finite Element Method (FEM)~\cite{reddy1993introduction} discretizes the continuous object, allowing the construction of global mass ($M$) and stiffness ($K$) matrices, which encapsulate the inertial and elastic properties of the object, respectively. Assuming small deformations and linear elastic material, the undamped free vibration of the discretized object is governed by~\cite{reddy1993introduction}:
\begin{equation}
M \ddot{x} + K x = 0,
\end{equation}
where $x$ is the displacement vector of the mesh nodes, and $\ddot{x}$ represents nodal accelerations. By assuming a set of harmonic solutions of the form $x_j = u_j e^{i\omega_j t}$, where $u_j$ is the mode shape and $\omega_j$ is the $j-$th angular natural frequency, it transforms into a generalized eigenvalue problem as:
\begin{equation}
K U =  M U \Lambda.
\end{equation}
Here, $\Lambda$ is a diagonal matrix of the object's eigenvalues $\lambda_1, \dots, \lambda_n$, where $\lambda_j = \omega_j^2$, and $U = [u_1, \dots, u_n]$ contains mode shapes (eigenvectors) of the vibrations. Solving this problem yields a set of eigenvalues $\lambda_j$, which relate to the natural frequencies of the object's vibration ($f_j = \frac{\sqrt{\lambda_j}}{2\pi}$), and their corresponding mode shapes (eigenvectors) $u_j$. 

\paragraph{Sound Synthesis.}

Once the eigenvalues $\lambda_i$ and eigenvectors $u_i$ are computed, The nodal displacement $x(t)$ can be represented as a linear combination of its mode shapes and reduced modal coordinates $q_i(t)$~\cite{barbicsiggraph}:
\begin{equation}
x(t) = \sum_{i} u_i q_i(t).
\end{equation}
For the full damped equation of motion~\cite{chowdhury2003computation}:

\begin{equation}
M \ddot{x} + C \dot{x} + K x = f(t), \label{DampedEM}
\end{equation}

where $C = \alpha M + \beta K$ is the Rayleigh damping matrix,
we can decouple \cref{DampedEM} for the $i$-th mode using $q_i(t)$:
\begin{equation}
\ddot{q}_i(t) + (\alpha + \beta \lambda) \dot{q}_i(t) + \lambda_i^2 q_i(t) = F_i(t) \label{Ft}
\end{equation}
where $F_i(t)=u_i^Tf(t)$ is the modal force of $i-$th mode reduced from nodal force $f(t)$, and $\zeta_i$ is the modal damping ratio.
For an impulse excitation $F_i(t)$ at $t=0$, the solution to each mode is a damped sinusoid. The resulting audio signal is their superposition:
\begin{equation}
S(t) = \sum_{i} A_i e^{-\sigma_i t} \sin(\omega_{d,i} t)\label{St}
\end{equation}
where $A_i$ is the amplitude of mode $i$, $\omega_{d,i}$ is the damped natural frequency, and $\sigma_i$ is the decay rate of mode $i$.

\subsection{Dataset Components and Formulation}

Formally, each sample in the VibraVerse dataset consists of the following components:
\begin{itemize}
    \item \textbf{3D Geometry}: Watertight 3D models defined by triangle surface meshes and their corresponding tetrahedral volumetric discretizations.
    \item \textbf{Visual Representation}: A single-view image rendered from a fixed viewpoint, using predefined materials.
    \item \textbf{Material Properties}: Physical attributes such as density ($\rho$), Young's modulus ($E$), and Poisson's ratio ($\nu$) that define the object's material characteristics.
    \item \textbf{Modal Parameters}: Eigenfrequencies and modal shapes obtained through modal analysis, which describe the object's vibrational behaviors.
    \item \textbf{Audios}: Sample audio signals generated based on the modal parameters and the Rayleigh damping coefficient (alpha, beta).
    \item \textbf{Metadata}: Additional information such as object category, size, source, and other relevant attributes.
\end{itemize}

\subsection{VibraVerse Dataset Creation Pipeline}

\begin{figure*}[htbp]
  \centering
  \includegraphics[width=0.95\linewidth]{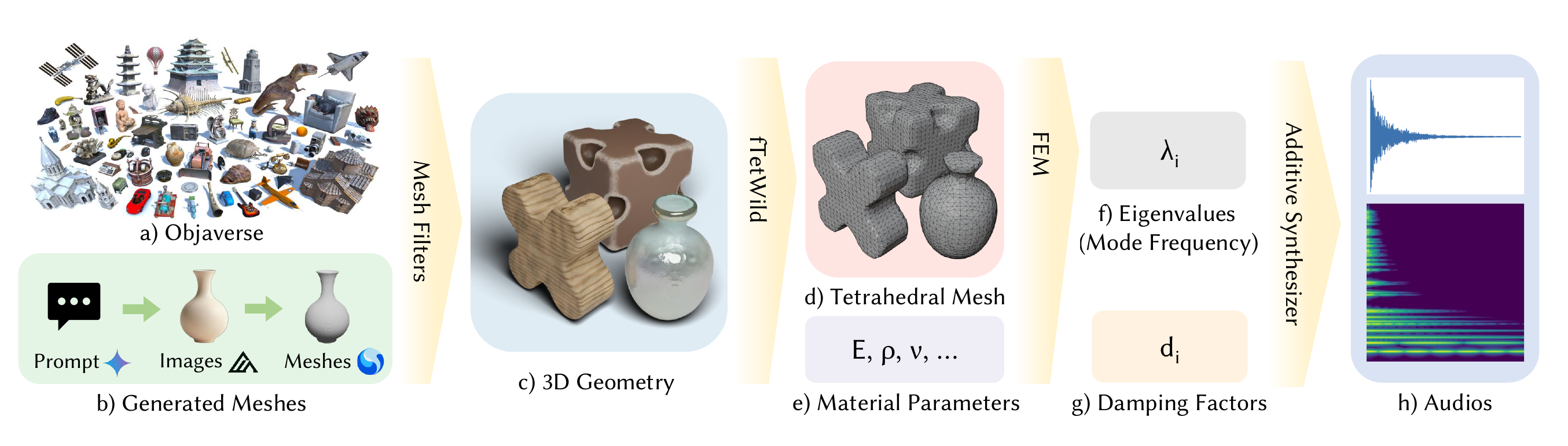}
  \vspace{-0.1in}
  \caption{Pipeline for generating our VibraVerse dataset. Meshes from Objaverse and text-to-3D generation are filtered and then tetrahedralized, assigned material parameters, and analyzed via finite-element modal analysis to obtain eigenvalues and damping factors. An additive synthesizer then produces corresponding modal sounds, forming physically consistent geometry–acoustics pairs.}
  \vspace{-0.1in}
  \label{fig:overall_pipeline}
\end{figure*}

\subsubsection{Geometry Creation, Processing}

We source 3D geometries from a combination of publicly available datasets and procedural generative techniques. 

\paragraph{Geometry Sources Part I: Objaverse.} 
We curated an open-source dataset from Objaverse~\cite{deitke2023objaverse} by applying a rigorous filtering process based on Objaverse++~\cite{lin2025objaverse++} annotations. The selection criteria required models to be single, non-scene, non-transparent, non-humanoid objects with a quality score $\geq 2$ and a file size $<5MB$.

\paragraph{Geometry Sources Part II: Generated.}
We generated a synthetic dataset of 40,000 models using a two-stage pipeline. First, Flux Dev~\cite{labs2025flux1kontextflowmatching} synthesized 2D images from prompts generated by Google Gemini~\cite{team2024gemini}. Second, Hunyuan3D 2.0~\cite{zhao2025hunyuan3d} reconstructed 3D geometries from these images. This process yielded 2,000 instances for each of 20 distinct categories (see Appendix for details).

\paragraph{Geometry Processing.}
The acquired models often contain simulation-inhibiting defects (e.g., holes, non-manifold edges). Our preprocessing pipeline first normalizes all models (translation to origin, scaling to [-1, 1]). We then perform a voxel remesh to generate high-quality watertight manifolds from the (potentially non-watertight) inputs. Finally, the mesh is made simulation-friendly using the method from GeGnn~\cite{pang2023}.

\subsubsection{Physical Validity Filtering}

\noindent
To ensure both the physical validity and representational diversity of the dataset, we implement a rigorous multi-stage geometry filtering and validation pipeline prior to modal analysis and acoustic synthesis.  
This process enforces structural coherence, topological simplicity, and physical plausibility, thereby ensuring that all retained models serve as stable and meaningful samples for geometry-acoustics learning. Specifically:

    \textbf{Topological Connectivity Filtering.}  
    We first eliminate geometries that contain multiple disconnected components or floating elements, which violate the assumption of a single physically coherent body required for modal analysis. Connectivity is assessed using graph-based component labeling and surface adjacency detection. Only single-connected
    meshes are preserved to guarantee well-defined boundary conditions for vibration mode computation. 

    \textbf{Topological and Geometric Complexity Control.}
    To prevent numerical instability and the overrepresentation of degenerate structures, we exclude 
    non-manifold geometries, and models exceeding a specified topological genus threshold are removed.  
    This ensures the dataset remains computationally tractable and physically interpretable.

    \textbf{Physical Plausibility and Modal Validity Screening.}
    Finally, each candidate undergoes a physical sanity check to validate its suitability for finite-element modal analysis.  
    We discard meshes that fail to satisfy minimum thickness constraints (to avoid thin-shell structures prone to non-linear vibrations) or that yield numerically unstable or non-physical eigenvalue spectra, such as negative eigenvalues, to preserve the integrity of the geometry–physics correspondence.

Through this hierarchical filtering pipeline, the resulting dataset achieves a high level of geometric fidelity, topological soundness, and physical realism, providing a reliable foundation for downstream 
physically-consistent learning tasks.
This also helps maintain a high-quality dataset that is diverse and plentiful.
After the above filtering steps, we obtain a final set of approximately 46,000 high-quality 3D geometries for inclusion in the VibraVerse dataset, 10,000 from Objaverse++ and 36,000 from our generation pipeline.

\subsection{Material Properties}
The material properties of each object are crucial for determining its vibrational characteristics and the resulting sound. We assign material properties based on a predefined set of common materials, such as wood, metal, plastic, and glass. Each material is characterized by its density $\rho$, Young's modulus $E$, and Poisson's ratio $\nu$.
Specifically, we define a material library with the following materials and their corresponding properties (See Appendix for details). The dataset contains 10 material categories: wood, plastic, ceramic, glass, steel, copper, aluminum, concrete, stone, and polycarbonate.
To determine the material properties for each object, we employed two distinct strategies. For models originating from the Objaverse dataset, we leveraged a Vision-Language Model (VLM) to classify their rendered visual appearance into a set of predefined material categories. For our procedurally generated models, we programmatically assigned a plausible material category based on the object's semantic class.

\subsection{Sound Synthesis}

For each 3D object with its geometry and material properties defined, we compute its natural frequencies, which are subsequently used to synthesize corresponding impact sounds.
The process is as follows: first, we convert the 3D geometry into an explicit volumetric tetrahedral mesh using the method in fTetWild~\cite{hu2020fast}. Subsequently, following the Modal Analysis method detailed in Section 3.1, we solve the 64 smallest eigenvalues $\lambda_i$, which correspond to the squared natural frequencies of the first 64 vibrational modes $\omega_i = \sqrt{\lambda_i}$. This decomposition is performed using the ARPACK library~\cite{lehoucq1998arpack}. 

Finally, to synthesize the impulse audio waveform, we apply a unit impulse excitation $\delta(t)$ to each mode:
\begin{equation}
\label{eq:impact}
F_i(t) = \delta(t), i=1,\dots,64
\end{equation}
Substituting this into \cref{Ft} allows us to solve for the amplitude $A_i$, damped frequency $\omega_i$ and damping coefficient $d_i$ of each mode's time-dependent vibration signal:
\begin{equation}
S_i(t) = A_i e^{-d_i t}sin(2\pi \omega_i t). \label{Sit}
\end{equation}
Following \cref{Sit}, we sample a 1-second signal at a sample rate of 32,000 Hz for each of the 64 modes. The resulting signals $S(t)$ are then summed to produce the object's corresponding impact sound.




 \section{Benchmark Tasks and Validation}

We designed several benchmark tasks to validate VibraVerse for physically-consistent multimodal learning. These tasks evaluate the cross-modal mappings between 3D geometry, materials, and acoustics, covering applications like conditional generation, reconstruction, and retrieval. We detail the experimental setup, methods, evaluation protocols, and quantitative/qualitative results for each task to demonstrate the dataset's effectiveness.


Unless otherwise specified, the following experiments are all based on our full dataset, with a training/testing split of 90\%/10\%. We provide the technical detail of each task and experimental settings in the supplementary material.

\subsection{Geometry$\rightarrow$Sound: Data-Driven Synthesis}

Given the geometry and material parameters of a 3D object, FEM-based modal analysis shows the process of synthesizing its impact sound (\cref{sec:method_background}). 
However, it requires generalized eigenvalue decomposition on large matrices, which is computationally expensive and slow. 
To evaluate the effectiveness of the VibraVerse dataset for data-driven sound synthesis, we design a learning-based task in which a neural network takes an object's 3D geometry and material parameters (density, Young’s modulus, and Poisson’s ratio) as input and directly predicts its natural frequencies.



We use an OCNN-based~\cite{Wang2017} shape encoder to extract geometric features from the input 3D shapes, and a multi-layer perceptron (MLP) is used to encode the material parameters,
then both features are concatenated and fed into a Sinusoidal Representation Network (SIREN~\cite{Sitzmann2020}), which 
predicted the first 64 modal frequencies. 
We minimize the mean squared error (MSE) between the predicted scaled frequencies and their corresponding ground-truth values.

Specifically, we compare to the following methods:
\begin{itemize}
    \item \textbf{NeuralSound~\cite{jin2022neuralsound}:} Retraining NeuralSound on our dataset even improves its vibration-solver performance beyond the original report.
    \item \textbf{FEM~\cite{reddy1993introduction}}: We perform FEM-based modal analysis using two eigenvalue solvers:ARPACK and LOBPCG.
\end{itemize}

We evaluate the quality and efficiency of all methods using two metrics. The Frequency error is defined as the Mean Squared Error (MSE) between predicted and ground-truth frequencies on scaled Mel spectrograms. The Time cost is measured as the total time taken to process the test set, which contains approximately 4,600 meshes. The detailed statistics are shown in \cref{tab:comparison_SS}.

\begin{table}[h!]
\centering
\caption{Comparison of different methods. Note that ARPACK and LOBPCG, being traditional FEM-based solvers, are generally treated as ground truth. Our dataset enables superior performance.}
\label{tab:comparison_SS}
\begin{tabular}{lcc}
\toprule
\textbf{Method} & \textbf{Freq. error $\downarrow$ } & \textbf{Time cost (s) $\downarrow$ } \\
\midrule
FEM (ARPACK) &  $<10^{-7}$ & 15374 \\
FEM (LOBPCG) & $<10^{-7}$ & 14112 \\
NeuralSound & $3.50 \times 10^{-3}$ & 441 \\
Ours & $6.06 \times 10^{-4}$ & \textbf{353} \\
\bottomrule
\end{tabular}
\end{table}

\subsection{Sound-Guided Shape Reconstruction}

\begin{figure}[htbp]
  \centering
  \includegraphics[width=0.88\linewidth]{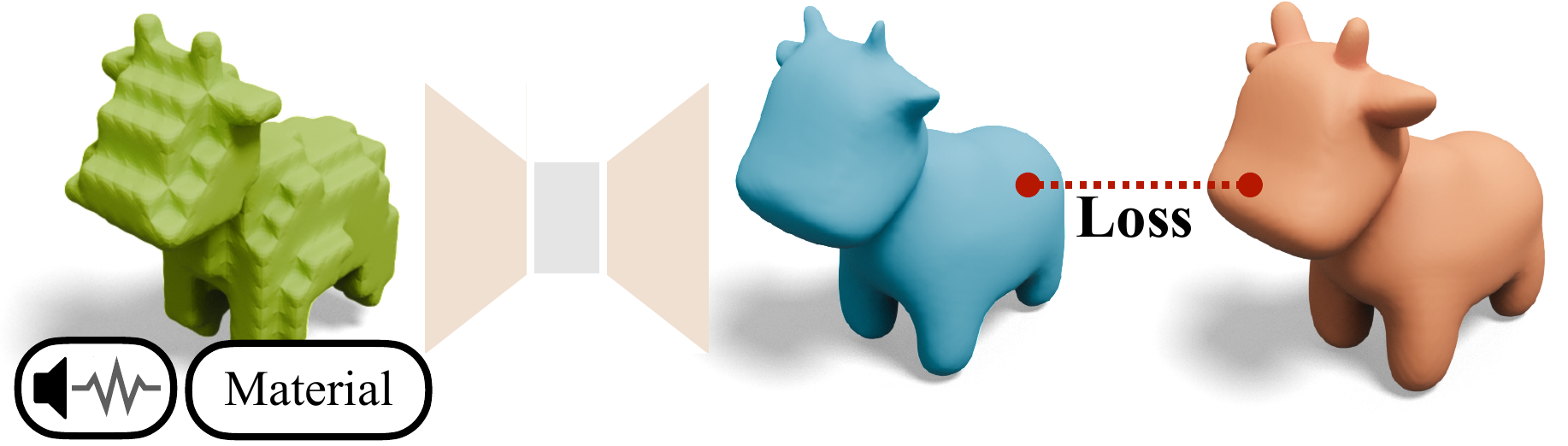}
  \caption{Sound-Guided Shape Reconstruction. Given a voxel initial shape, the audio eigenvalues, and material properties, we reconstruct the 3D geometry in just one forward pass.}
  \label{fig:sound_guided_shape_reconstruction_pipeline}
\end{figure}


Inferring a complete 3D geometry from impact sound is inherently an ill-posed problem: acoustic responses encode only a subset of an object’s modal characteristics and are highly sensitive to local structural variations, allowing distinct geometries to produce nearly indistinguishable sound patterns.
To date, the only method capable of performing geometry-from-sound inversion is DiffSound~\cite{Jin2024DiffSound}, which introduces a differentiable, physics-based modal sound rendering framework to infer geometry under a coarse voxel constraint.
Motivated by this challenge, we designate sound-guided shape reconstruction as one of the core benchmark tasks in VibraVerse, enabling the systematic evaluation of learnable and generalizable geometry inference from sound.

The overall pipeline of this reconstruction is illustrated in ~\cref{fig:sound_guided_shape_reconstruction_pipeline}.
Same as ~\cite{Jin2024DiffSound}, we take sound eigenvalues, as well as a coarse voxel grid and material parameters as input, and reconstruct the detailed 3D geometry.
We apply the VAE structure and training methodology of Step1X-3D~\cite{li2025step1x}. We use the VAE to encode the voxel grid, which is then concatenated with the encoded audio features and material embeddings as conditions. The combined features are then fed into a decoder network to reconstruct the final 3D geometry.
We compared the performance of DiffSound and our method on this task. 
We randomly sampled 100 test meshes from the test set, evaluated the accuracy using Intersection over Union (IoU) and Chamfer Distance (CD), and measured efficiency using inference time on test meshes.
Quantitative results are reported below, and qualitative examples are shown in~\cref{fig:sound_guided_shape_reconstruction_visual_result}.
The results show that our VibraVerse dataset can facilitate a data-driven approach to directly reconstruct 3D geometry from audio, achieving simultaneous improvements in both accuracy and efficiency.





\vspace{0.5em}

\label{tab:comparison_SR}
\hspace*{-0.4cm}
\begin{tabular}{lccc} 
\toprule
\textbf{Method} & \textbf{IoU} $\uparrow$ & \textbf{CD} $\downarrow$ & \textbf{Time(s)} $\downarrow$ \\
\midrule
Initial Voxel &  0.837 & $4.97 \times 10^{-3}$   & \textbackslash \\
DiffSound & 0.856 & $4.45 \times 10^{-3}$ & 46594 \\
Ours & \textbf{0.871} &  $\boldsymbol{3.32 \times 10^{-3}}$ & \textbf{175} \cyan{($\times 0.004$)} \\
\bottomrule
\end{tabular}

\begin{figure}[htbp]
  \centering 
  \includegraphics[width=0.8\linewidth]{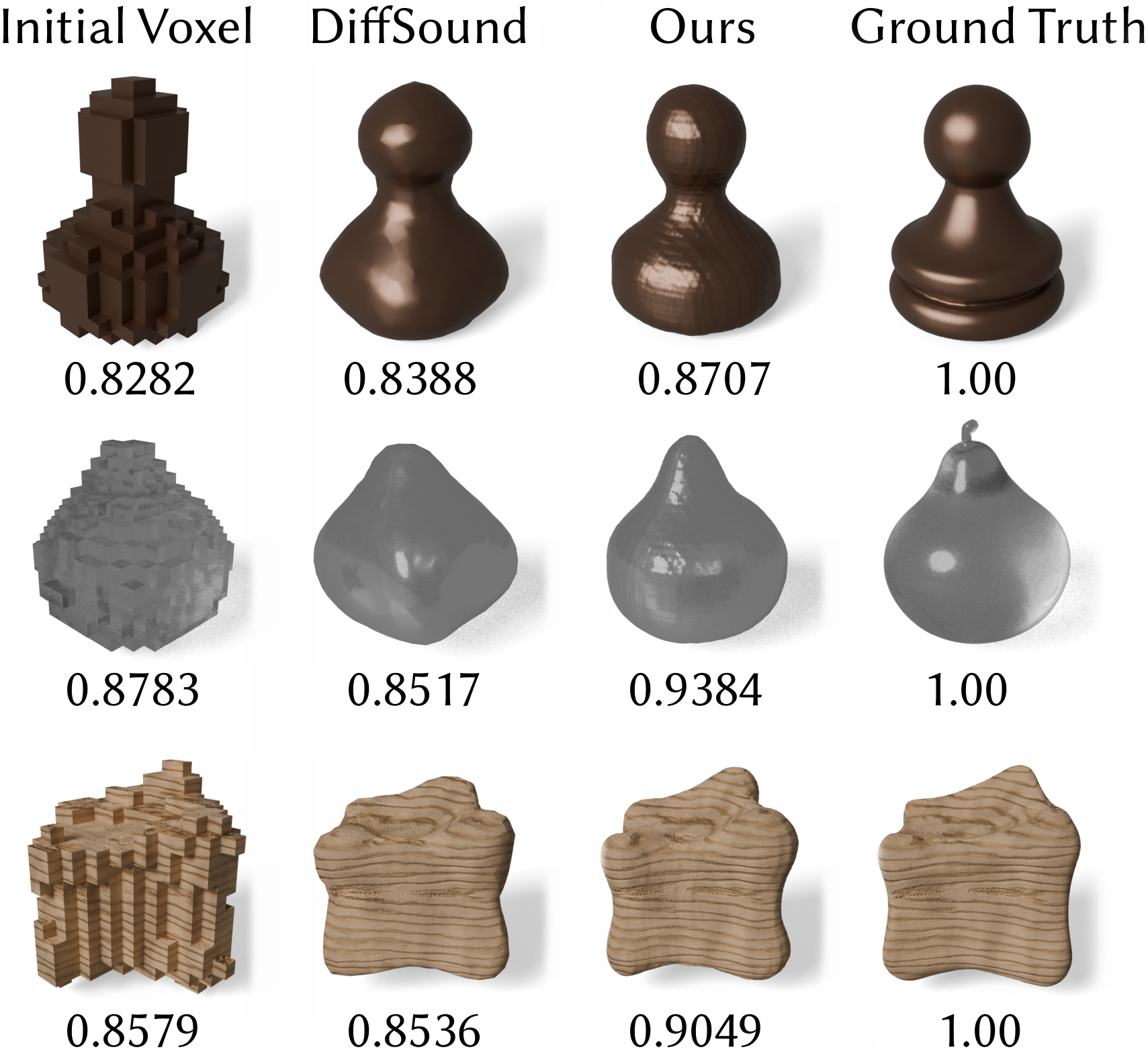}
  \caption{Results of audio-guided reconstruction. From left to right are initial shapes, DiffSound results, our results, and the ground truth. The IoU metric is shown below each shape.}
  \label{fig:sound_guided_shape_reconstruction_visual_result}
\end{figure}

\subsection{Cross-Modal Retrieval} 
\label{sec:cross_modal_retrieval} 

\begin{figure}[htbp]
  \centering
  \includegraphics[width=0.8\linewidth]{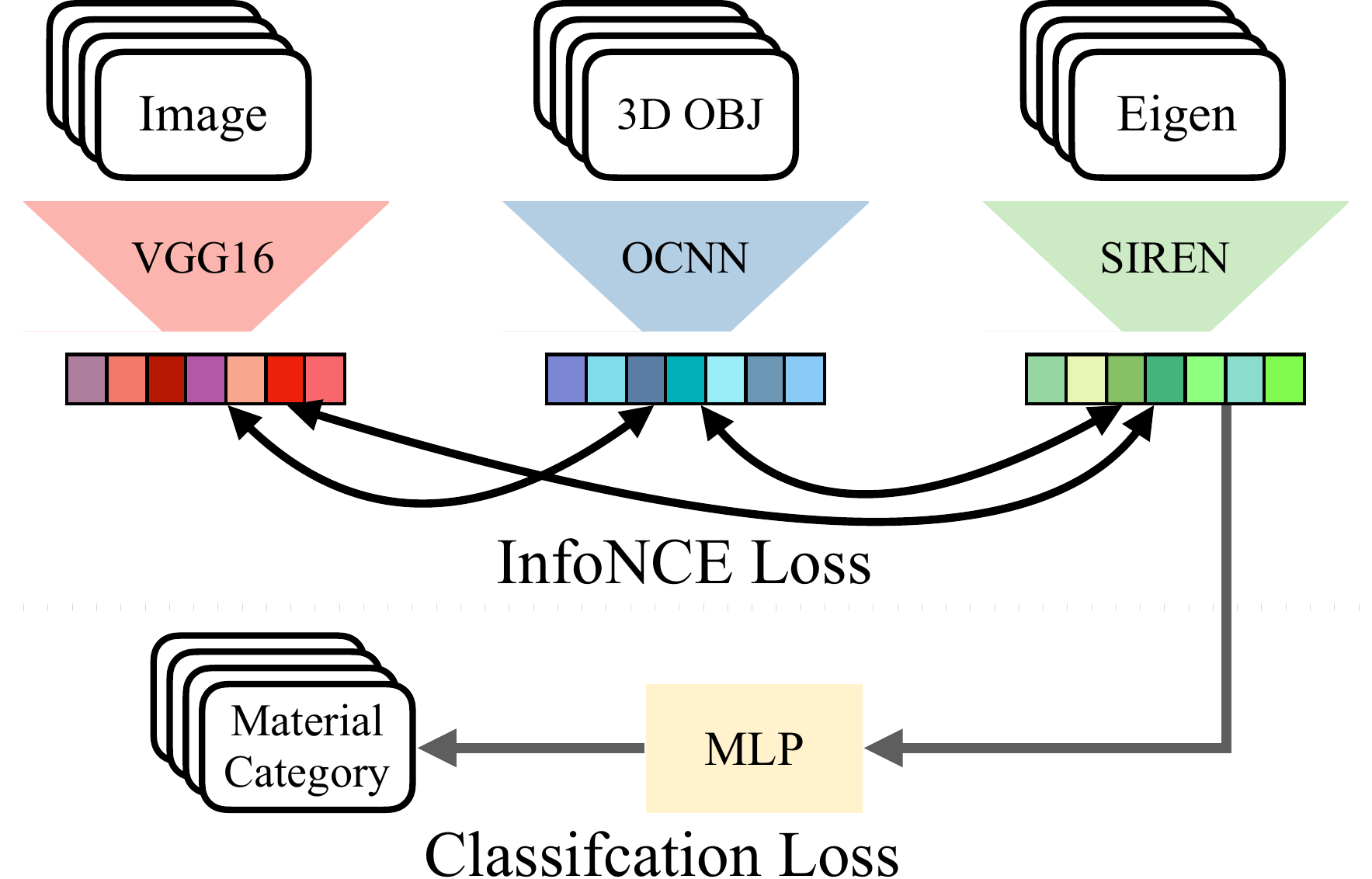}
  \caption{Architecture of the CLASP model. Three encoders are used to extract features from each modality, and a contrastive learning mechanism is employed to align the features.}
  \label{fig:clasp_architecture}
\end{figure}

\begin{figure*}[htbp]
  \centering
  \includegraphics[width=0.85\linewidth]{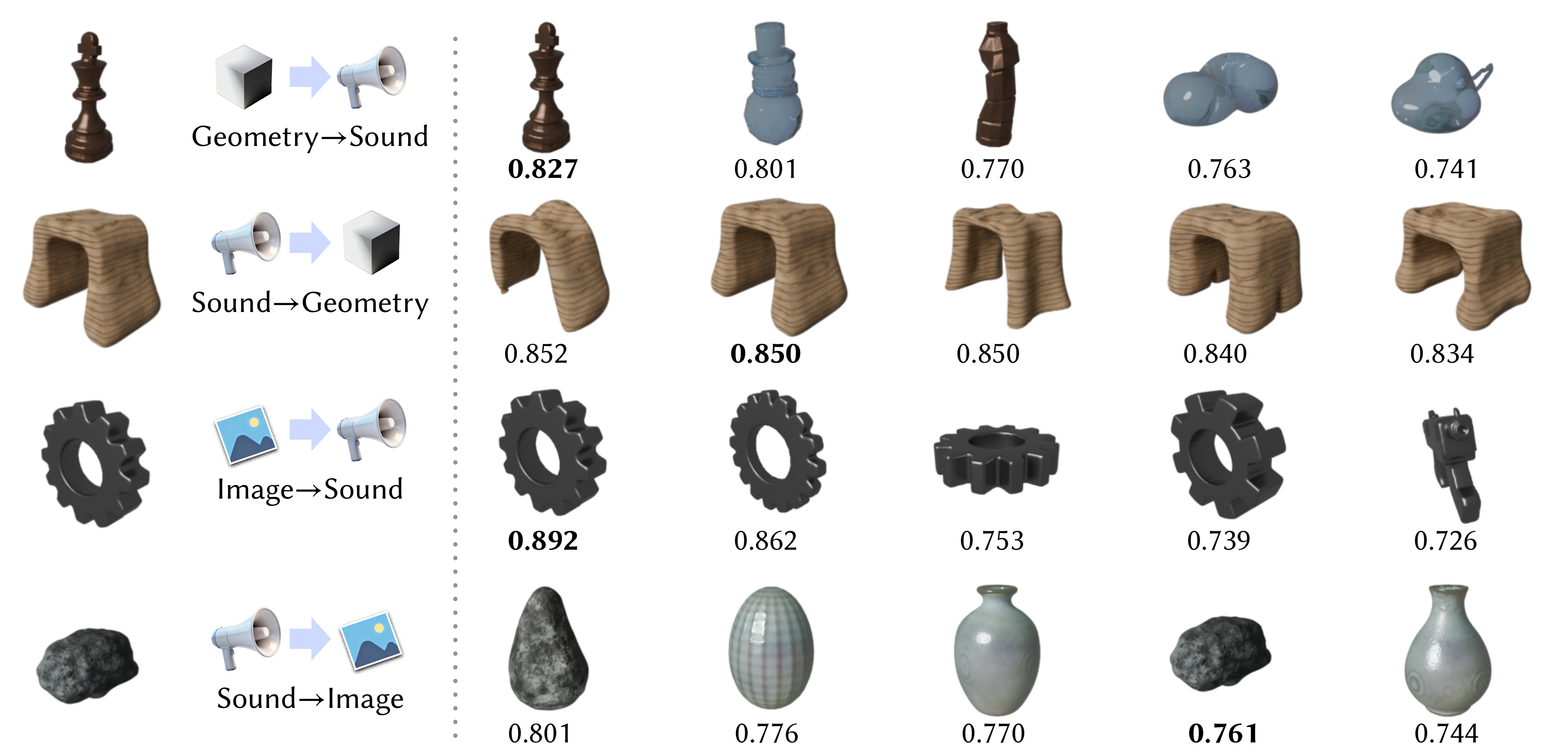}
  \caption{Cross-modal retrieval results. Given a query from one modality (left), we retrieve the most relevant items from other modalities (right). The numbers are the cosine similarity, while the bold number indicates the ground truth. 
  }
  \label{fig:retrieval_samples}
\end{figure*}

\begin{table*}[ht]
\centering

\label{tab:retrieval_performance}
\begin{tabular}{lccccccccc}
\toprule
& \multicolumn{3}{c}{\textbf{Overall (\#Sample = 4672)}} & \multicolumn{3}{c}{\textbf{Objaverse (\#Sample = 1117)}} & \multicolumn{3}{c}{\textbf{Generated (\#Sample = 3555)}} \\
\cmidrule(lr){2-4} \cmidrule(lr){5-7} \cmidrule(lr){8-10}
\textbf{Task} & {\textbf{R@1}} & {\textbf{R@5}} & {\textbf{R@10}} & {\textbf{R@1}} & {\textbf{R@5}} & {\textbf{R@10}} & {\textbf{R@1}} & {\textbf{R@5}} & {\textbf{R@10}} \\
\midrule
Geometry $\rightarrow$ Sound & 0.409 & 0.766 & 0.865 & 0.288 & 0.654 & 0.787 & 0.484 & 0.848 & 0.929 \\
Sound $\rightarrow$ Geometry & 0.417 & 0.768 & 0.866 & 0.312 & 0.662 & 0.797 & 0.488 & 0.854 & 0.936 \\
Image $\rightarrow$ Sound    & 0.287 & 0.610 & 0.727 & 0.207 & 0.475 & 0.604 & 0.334 & 0.688 & 0.807 \\
Sound $\rightarrow$ Image    & 0.308 & 0.617 & 0.732 & 0.224 & 0.474 & 0.610 & 0.359 & 0.702 & 0.818 \\
Geometry $\rightarrow$ Image & 0.509 & 0.837 & 0.909 & 0.474 & 0.815 & 0.892 & 0.548 & 0.873 & 0.930 \\
Image $\rightarrow$ Geometry & 0.499 & 0.835 & 0.914 & 0.471 & 0.806 & 0.895 & 0.532 & 0.869 & 0.936 \\
\bottomrule
\end{tabular}
\caption{
  Cross-modal retrieval performance (R@1, R@5, R@10) of our datasets.
  Retrieval between geometry and sound achieves higher accuracy compared to retrieval between image and sound, likely due to the more direct correlation between geometry and sound.
  Our result has strong performance on all retrieval tasks, demonstrating the effectiveness of our VibraVerse dataset in facilitating cross-modal learning.
}
\end{table*}

Our VibraVerse dataset provides a unified platform to explore the mutual correspondence between shape, vision, and sound.
To demonstrate the effectiveness of our dataset in terms of cross-modal tasks, inspired by CLIP~\cite{Radford2021a}, we design a contrastive learning~\cite{chen2020simple,chen2020big} framework for cross-modal retrieval between 3D shapes, 2D images, and sounds, named \textit{Contrastive Learning of Audio, Shape, and Physical-Properties} (CLASP), as in~\cref{fig:clasp_architecture}.
We use different encoders to extract features from each modality, and train the model using a contrastive loss to align the embeddings of matching pairs while pushing apart non-matching pairs, with the InfoNCE loss:
\begin{equation}
\mathcal{L} = - \log \frac{\exp(\text{sim}(\mathbf{z}_i, \mathbf{z}_j)/\tau)}{\sum_{k=1}^{N} \exp(\text{sim}(\mathbf{z}_i, \mathbf{z}_k)/\tau)} \ ,
\end{equation}
where $\mathbf{z}_i$ and $\mathbf{z}_j$ are the embeddings of a matching pair from different modalities, $\text{sim}(\cdot, \cdot)$ denotes the cosine similarity, $\tau$ is a temperature hyperparameter set to 0.07, and $N$ is the total number of samples in the batch.

The model consists of a SIREN-based ~\cite{Sitzmann2020} sound encoder, an OCNN-based ~\cite{Wang2017} 3D shape encoder, and a VGG-based ~\cite{simonyan2014very} image encoder.
To retrieve the most relevant item from a library, we compute the cosine similarity between the query embedding and all candidate embeddings, selecting the those with highest similarity scores.


We consider cross-modal retrieval tasks across three modalities: audio (or its eigenvalues), 3D geometry, and 2D images. The tasks focus on the bidirectional retrieval between sound and 3D shapes, and between sound and 2D images.
The visual result of cross-modal retrieval is shown in ~\cref{fig:retrieval_samples}. 
Quantitative results are presented in \cref{tab:retrieval_performance}, where we report the Recall@K (R@K) metrics for each retrieval task across different subsets of our VibraVerse dataset.
A more detailed comparison between our dataset and Objectfolder~\cite{deitke2023objaverse} is provided in the supplementary material.

\subsection{Sound$\rightarrow$Material: Material Classification}
\label{sec:material_pred}
Material classification is defined as predicting an object's material category solely from its acoustic properties. 
We add a classification head in CLASP, as the bottom part of ~\cref{fig:clasp_architecture}, which takes in the embedding extracted from encoder and predicts the material category. The prediction accuracies are reported below, which suggest that our dataset effectively supports material classification tasks.

\vspace{0.5em}

     \begin{tabular}{lccc}
     \toprule
            & Objaverse & Generated & All \\  \midrule
         Accuracy $\uparrow$& 51.03\% & 89.54\% & 80.33\% \\ 
     \bottomrule 
     \end{tabular}



\vspace{0.5em}

Some previous works ~\cite{clarke2023realimpact,gao2021ObjectFolder,gao2022ObjectFolderV2,Gao_2023_CVPR} have also explored this task, but with different settings. For a detailed comparison, please refer to our supplementary material.

\subsection{Sound-Guided Solid Identification}
\label{sup:solid}
Visual modalities alone lack the information density required to infer internal physical properties, such as differentiating between solid and hollow structures. Impact sounds, however, offer extrinsic evidence of these internal characteristics. As a supplementary experiment, we formulate a binary classification task that leverages both single-view images and the modal frequencies of impact sound to identify structural solidity.

\noindent
\textbf{Training Data.} 
Adopting the hollow mesh generation methodology proposed in DiffSound \cite{Jin2024DiffSound}, we generate hollow counterparts of existing solid objects by removing their interiors.

Specifically, according to \cite{Jin2024DiffSound}, we define the "thickness" of generated hollow mesh based on the Signed Distance Function (SDF) of its corresponding solid counterpart. Let $s_{\min}$ denote the global minimum SDF value, corresponding to the internal point strictly furthest from the surface (i.e., the maximum depth). We define a hollow object with a relative thickness ratio $t$ as the set containing all points whose distance to the surface is within $t \cdot |s_{\min}|$. Consequently, a point $P$ is considered to be inside the hollow shell if and only if its SDF value, $\mathcal{S}(P)$, satisfies the condition:
\begin{equation}
    t \cdot s_{\min} < \mathcal{S}(P) < 0
\end{equation}
For our dataset creation, we synthesize hollow meshes by uniformly sampling the thickness ratio $t$ from the range $[0.3, 0.7]$.
Crucially, this process preserves the exterior mesh, rendering the hollow and solid objects visually indistinguishable.

We synthesized 1,000 hollow objects alongside their modal frequencies. By combining these with the existing solid counterparts from the original dataset, we constructed a balanced dataset of 2,000 samples. Each data entry comprises multi-modal inputs (audio and image) and a binary label (solid or hollow). Finally, the dataset was partitioned into a training set of 1,600 samples and a test set of 400 samples.

\noindent
\textbf{Methods.} 
Following the design in \cref{sec:cross_modal_retrieval}, we employ a SIREN-based~\cite{Sitzmann2020} sound encoder and a VGG-based~\cite{simonyan2014very} image encoder. Specifically, the input image and modal frequencies are processed by their respective encoders to extract visual and auditory embeddings. These feature vectors are subsequently concatenated and passed through a Multi-Layer Perceptron (MLP) to generate a two-dimensional output, representing the logits for the solid and hollow classes. The model is optimized using cross-entropy loss. We train the model for 100 epochs in training set, which takes approximately 3 hours on a NVIDIA RTX 4090 GPU.

\begin{figure}[thbp]
  \centering 
  \includegraphics[width=1.0\linewidth]{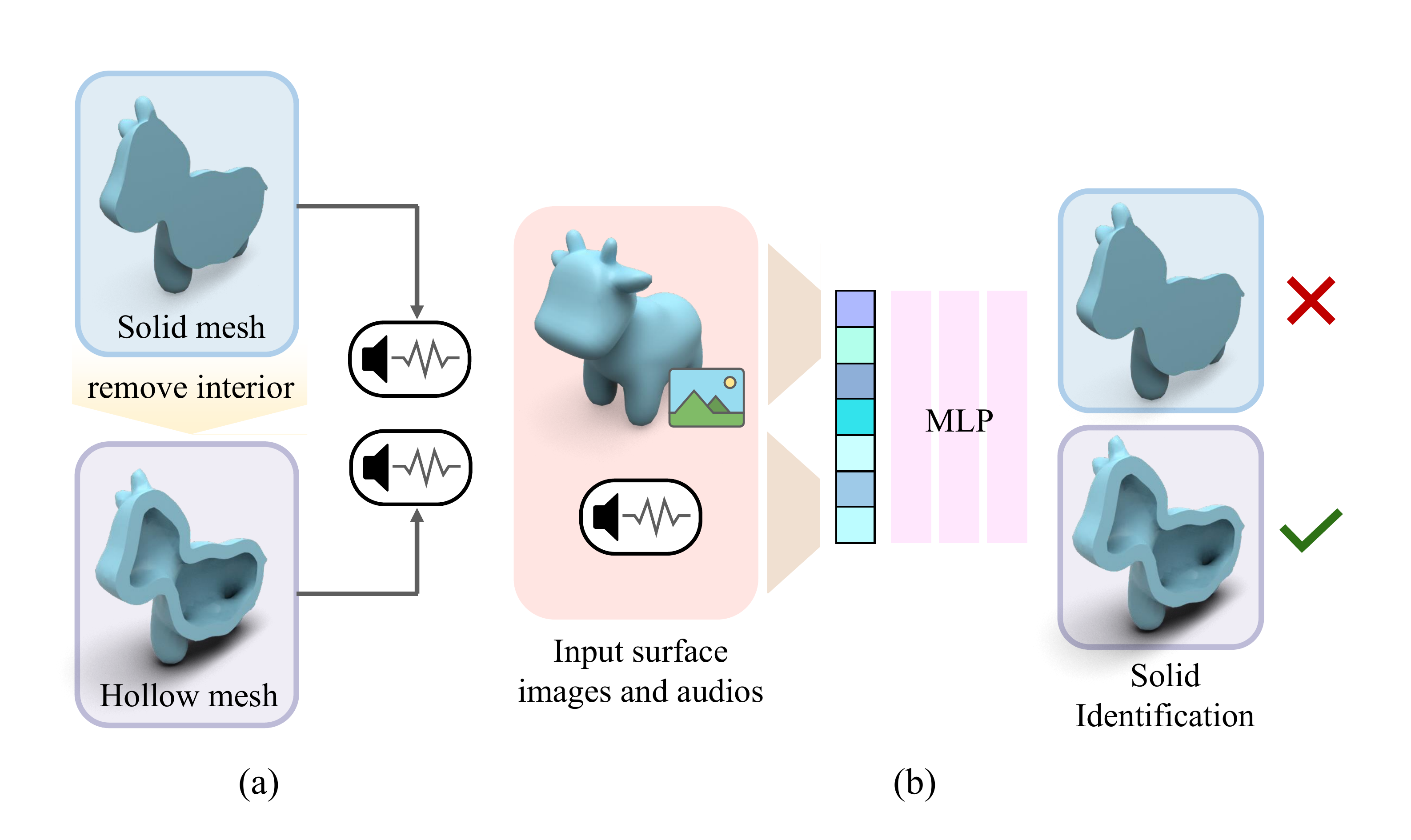}
  \caption{Sound-Guided Solid Identification. (a) For solid objects, we construct hollow counterparts by removing their interior. We then synthesize audios for both the solid and hollow objects. (b) Taking the surface rendering and modal frequencies as inputs, the model classifies whether the source object is solid or hollow.}
  \label{fig:sound_guided_shape_reconstruction_visual_result}
\end{figure}

\noindent
\textbf{Experiment Results.} 
 The classification accuracy on the test set is presented below. Furthermore, we evaluate the performance separately on the two distinct data sources: Objaverse and Generated items. The results demonstrate that our dataset enables data-driven approaches to effectively recognize internal object structures by leveraging audio cues.

\vspace{0.5em}
\begin{tabular}{lccc}
     \toprule
            & Objaverse & Generated & All \\  \midrule
         Accuracy $\uparrow$& 74.00\% & 91.33\% & 87.00\% \\ 
     \bottomrule 
     \end{tabular}

 \section{Conclusion}

Our VibraVerse dataset and benchmark suite provide a large-scale, physics-grounded foundation in which geometry, material, and sound are explicitly coupled through physically consistent simulation.
By integrating principles from computational acoustics into multimodal learning, it enables models to infer geometric and material properties from auditory cues, marking a step toward physically interpretable auditory intelligence.

We reinterpret modal analysis as a language of physical behavior that reveals how objects vibrate, store energy, and express their intrinsic properties. As the dynamic fingerprint of an object, it provides the causal foundation linking geometry, material, and sound within a unified representation framework.
Building on this perspective, our work advances sound-guided 3D perception, physics-consistent multimodal reasoning, and embodied physical understanding, with promising potential for sim-to-real transfer.

Nevertheless, our approach has limitations. All data in VibraVerse are fully synthetic and generated under idealized simulation conditions, which may not fully capture the noise and variability of real-world acoustic measurements. Moreover, the benchmark has not yet been validated against real recordings or experimentally measured modal properties, leaving the degree of sim-to-real generalization to be explored in our future work.

Beyond serving as a benchmark for multimodal reasoning, our dataset also holds potential for advancing physics-informed neural networks (PINNs) \cite{raissi2019physics} and neural-based physical simulation \cite{sanchez2020learning}, offering a pathway to unify data-driven learning with physically grounded modeling. These need further validation and will be our future work.


{
    \small
    \bibliographystyle{ieeenat_fullname}
    \bibliography{main}
}

\end{document}